\documentclass[preprint,12pt,authoryear]{elsarticle}

\usepackage{amssymb}

\usepackage{lipsum}
\usepackage{rotating}
\usepackage{tabularx}

\usepackage{makecell}
\usepackage{amsmath}

\DeclareMathOperator*{\argmin}{arg\,min}
\usepackage[hidelinks]{hyperref}

\usepackage{booktabs}

\usepackage{tikz}
\usetikzlibrary{shapes.geometric}
\usetikzlibrary{positioning,shapes,shadows,arrows,calc}
\usetikzlibrary{intersections}
\pgfdeclarelayer{background}
\pgfdeclarelayer{foreground}
\pgfsetlayers{background,main,foreground}

\usepackage{todonotes}
\usepackage{paralist}

\presetkeys{todonotes}{inline}{}

\newcommand{\inst}{i}
\newcommand{\insts}{\mathcal{I}}
\newcommand{\algo}{\mathbf{A}}
\newcommand{\portfolio}{\mathcal{P}}
\newcommand{\feat}{\mathcal{F}}

\newcommand{\data}{\mathcal{D}}
\newcommand{\x}{\mathbf{x}}

\journal{Artificial Intelligence}

\begin{document}

\begin{frontmatter}



\title{The Algorithm Selection Competitions 2015 and 2017}


\address[Freiburg]{University of Freiburg, Germany}
\address[Columbia]{Columbia University, USA}
\address[Wyoming]{University of Wyoming, USA}

\author[Freiburg]{Marius Lindauer}
\ead{lindauer@cs.uni-freiburg.de}

\author[Columbia]{Jan N. van Rijn}
\ead{j.n.vanrijn@columbia.edu}

\author[Wyoming]{Lars Kotthoff}
\ead{larsko@uwyo.edu}

\begin{abstract}
The algorithm selection problem is to choose the most suitable algorithm for
solving a given problem instance. It leverages the complementarity between
different approaches that is present in many areas of AI. We report on the state
of the art in algorithm selection, as defined by the Algorithm Selection
competitions in 2015 and 2017. The results of these competitions show
how the state of the art improved over the years. We show that although
performance in some cases is very good, there is still room for improvement in other
cases.
Finally, we provide insights into why some scenarios are hard, and pose challenges
to the community on how to advance the current state of the art.
\end{abstract}


\end{frontmatter}


\section{Introduction}

In many areas of AI, there are different algorithms to solve the same type of
problem. Often, these algorithms are complementary in the sense that one
algorithm works well when others fail and vice versa. For example in
propositional satisfiability solving (SAT), there are complete tree-based solvers
aimed at structured, industrial-like problems, and local search solvers aimed at
randomly generated problems. In many practical cases, the performance difference
between algorithms can be very large, for example as shown by \cite{xu-sat12a} for SAT.
Unfortunately, the correct selection of an algorithm
is not always as easy as described above and even easy decisions require substantial
expert knowledge about algorithms and the problem instances at hand.

Per-instance algorithm selection~\citep{rice76a} is a way to leverage this
complementarity between different algorithms. Instead of running a single
algorithm, a portfolio~\citep{huberman-science97a,gomes-aij01} consisting of
several complementary algorithms is employed together with a learned selector.
The selector automatically chooses the best algorithm from the portfolio for each instance to be solved.

Formally, the task is to select the best algorithm $\algo$ from a portfolio of
algorithms $\portfolio$ for a given instance $\inst$ from a set of
instances $\insts$ with respect to a performance metric $m: \portfolio \times
\insts \to \mathbb{R}$ (e.g., runtime, error, solution quality or accuracy). To
this end, an algorithm selection system learns a mapping from an instance to a selected algorithm
$s: \insts \to \portfolio$ such that the performance, measured as cost, across
all instances $\insts$ is minimized (w.l.o.g.):

\begin{equation}
\argmin_s \sum_{\inst \in \insts} m(s(\inst),\inst)\label{eq:as}
\end{equation}

Algorithm selection has gained prominence in many areas and made tremendous
progress in recent years. Algorithm selection systems established new state-of-the-art
performance in several areas of AI, for example propositional satisfiability
solving~\citep{xu-jair08a}\footnote{In propositional satisfiability solving
(SAT), algorithm selection system were even banned from the SAT competition for
some years, but are allowed in a special track now.}, machine
learning~\citep{Brazdil2008,Rijn2018}, maximum satisfiability
solving~\citep{ansotegui-aij16}, answer set
programming~\citep{lindauer-aij17a,calimeri2017dlv}, constraint
programming~\citep{hurley-cpaior14a,amadini-tplp14a}, and the traveling salesperson problem~\citep{kotthoff-lion15a}. However, the multitude of
different approaches and application domains makes it difficult to compare
different algorithm selection systems, which presented users with a very
practical meta-algorithm selection problem -- which algorithm 
selection system should be used for a given task.
The algorithm selection competitions can help users to make the decision which
system and approach to use, based on a fair comparison across a diverse range of
different domains.

The first step towards being able to perform such comparisons was the
introduction of the Algorithm Selection Benchmark Library
\citep[ASlib,][]{bischl-aij16a}.
ASlib consists of many algorithm selection scenarios for which
performance data of all algorithms on all instances is available.
These scenarios allow for fair and reproducible comparisons of different
algorithm selection systems.
ASlib enabled the
competitions we report on here.

\paragraph{Structure of the paper}
In this competition report, we summarize the results and insights gained by
running two algorithm selection competitions based on ASlib. These
competitions were organized in 2015 -- \emph{the ICON Challenge on Algorithm
Selection} -- and in 2017 -- \emph{the Open Algorithm Selection
Challenge}.\footnote{This paper builds upon individual short papers for each
competition~\citep{kotthoff-ai17a,Lindauer17} and presents a unified view with a
discussion of the setups, results and lessons learned.} We start
by giving a brief background on algorithm selection (Section~\ref{sec:back}) and
an overview on how we designed both competitions (Section~\ref{sec:setup}).
Afterwards we present the results of both competitions
(Section~\ref{sec:results}) and discuss the insights obtained and open
challenges in the field of algorithm selection, identified through the
competitions (Section~\ref{sec:insights}).

\section{Background on Algorithm Selection}
\label{sec:back}
In this section, we discuss the importance of algorithm selection,
several classes of algorithm selection methods and
ways to evaluate algorithm selection problems.

\subsection{Importance of Algorithm Selection}

The impact of algorithm selection in several AI fields is best 
illustrated by the performance of such approaches in AI competitions.
One of the first well-know algorithm selection systems was
SATzilla~\citep{xu-jair08a},
which won several first places in the SAT competition~2009 and the SAT challenge~2012.
To refocus on core SAT solvers, portfolio solvers (including algorithm selection systems)
were banned from the SAT competition for several years---now, they are allowed
in a special track.
In the answer set competition 2011, the algorithm selection system claspfolio~\citep{hoos-tplp14a} won
the NP-track and later in 2015, ME-ASP~\citep{maratea-jlc15a} won the competition.
In constraint programming, sunny-cp~\citep{amadini-tplp14a} won the open track of the MiniZinc Challenge for several years (2015, 2016 \& 2017).
In AI planning, a simple static portfolio of planners \citep[fast downward stone
soup;][]{helmert-icaps11a} won a track at the International Planning Competition
(IPC) in~2011.
More recently, the online algorithm selection system Delfi~\citep{katz-ipc18a} won a first place at IPC~2018.
In QBF, an algorithm selection system called QBF Portfolio~\citep{hoos-cp18a} won third place at the prenex track of QBFEVAL 2018.

Algorithm selection does not only perform well for combinatorial problems,
but it is also an important component in automated machine learning (AutoML) systems.
For example, the AutoML system auto-sklearn uses algorithm selection to initialize its
hyperparameter optimization~\citep{feurer-aaai15a} and won two AutoML challenges~\cite{feurer-automl18b}.

There are also applications of algorithm selection in non-AI domains, e.g.\
diagnosis~\citep{koitz_improving_2016}, databases~\citep{dutt_plan_2016}, and
network design~\citep{selvaraj_pce-based_2017}.

\subsection{Algorithm Selection Approaches}

\begin{figure}[t]
\centering
\tikzstyle{activity}=[rectangle, draw=black, rounded corners, text centered, text width=8em, fill=white, drop shadow]
\tikzstyle{data}=[rectangle, draw=black, text centered, fill=black!10, text width=8em, drop shadow]
\tikzstyle{myarrow}=[->, thick]
\begin{tikzpicture}[align=center,node distance=3.9cm]
  \node (Features) [activity] {Compute Features $\feat(\inst)$};
  \node (Instance) [data, left of=Features, text width=6em, xshift=0cm] {Instance $\inst$};
  \node (Select) [activity, right of=Features, text width=7.8em] {Select\\ $s(\feat(\inst)):=\algo \in \portfolio$};
  \node (Solve) [activity, right of=Select, text width=4em, xshift=-0.8cm] {Solve $\inst$\\ with $\algo$};
  \node (Portfolio) [data, above of=Select, yshift=-2.0cm] {Algorithm Portfolio $\portfolio$};

  \draw[myarrow] (Instance) -- (Features);
  \draw[myarrow] (Features) -- (Select);
  \draw[myarrow] (Select) -- (Solve);
  \draw[myarrow] (Portfolio) -- (Select);

\end{tikzpicture}
\caption{Per-instance algorithm selection workflow for a given instance.\label{fig:pias}}
\end{figure}
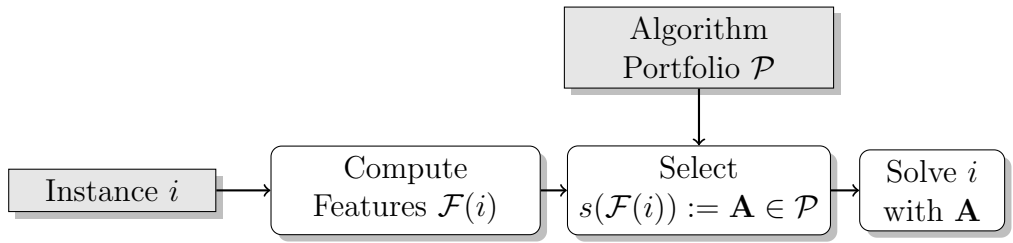

Figure~\ref{fig:pias} shows a basic per-instance algorithm selection framework
that is used in practice.
A basic approach involves
\begin{inparaenum}[(i)]
\item representing a given instance $\inst$ with a vector of numerical features $\feat(\inst)$
(e.g., number of variables and constraints of a CSP instance),
\item inducing a selection machine learning model $s$ that selects an algorithm for the given
instance $\inst$ based on its features $\feat(\inst)$.
\end{inparaenum}
Generally, these machine learning models are induced based on a dataset
$\data = \{(\x_j, y_j) \mid j = 1, \ldots, n\}$ with $n$ datapoints to map an
input $\x$ to output $f(\x)$, which closely represents $y$.
In this setting, $\x_i$ is typically the vector of numerical features $\feat(\inst)$ from some instance $\inst$ that has been observed before.
There are various variations for representing the $y$ values and ways for
algorithm selection system $s$ to leverage the predictions $f(\x)$.
We briefly review several classes of solutions.

\begin{description}
	\item[Regression] that models the performance of individual algorithms in the portfolio.
	A regression model $f_\algo$ can be trained for each $\algo \in \portfolio$
	on $\data$ with $\x_j = \feat(\inst)$ and $y_j = m(\algo, \inst)$ for each
	previously observed instance $i$ that $\algo$ was ran on.
	The machine learning algorithm can then predict how well algorithm $\algo$
	performs on a given instance from $\insts$.
	The algorithm with the best predicted performance is selected for solving
    the instance \citep[e.g.,][]{Horvitz2001,xu-jair08a}.
	\item[Combinations of unsupervised clustering and
	classification] that partitions instances into
    clusters $H$ based on the instance features $\feat(\inst)$,
	and determines the best algorithm $\algo_i$ for each cluster $h_i \in H$.
	Given a new instance $\inst'$, the instance features $\feat(\inst')$
    determine the nearest cluster $h'$ w.r.t. some distance metric; 
    the algorithm $\algo'$ assigned to $h'$ is applied
    \citep[e.g.,][]{ansotegui-cp09a}.
	\item[Pairwise Classification] that considers pairs
    of algorithms $(\algo_k,\algo_j)$. For a new instance, the machine-learning-induced model predicts for each pair of
    algorithms which one will perform better ($m(\algo_k,i) < m (\algo_j, i)$), and the algorithm with most ``is
    better'' predictions is selected \citep[e.g.,][]{xu-rcra11a,Rijn2015}.
	\item[Stacking of several approaches]
    that combine multiple models to predict the algorithm to choose, for example
    by predicting the performance of each portfolio algorithm through regression
    models and combining these predictions through a classification model
    \citep[e.g.,][]{kotthoff_hybrid_2012,samulowitz-sat13a,malone-mlj18a}.
\end{description}

\subsection{Why is algorithm selection more than traditional machine learning?}

In contrast to typical machine learning tasks,
each instance has a weight attached to it.
It is not be important to select the best algorithm on instances
on which all algorithms perform nearly equally, but
it is crucial to select the best algorithm on an instance
on which all but one algorithm perform poorly (e.g., all but one time out). The
potential gain from making the best decision can be seen as a weight for that
particular instance.

Instead of predicting a single algorithm, schedules of algorithms can also be used.
One variant of algorithm schedules~\citep{kadioglu-cp11a,hoos-tplp14b} are
static (instance-independent) pre-solving schedules which are applied before any
instance features are computed~\citep{xu-jair08a}. Computing the best-performing
schedule is usually an NP-hard problem.
Alternatively, a sequence of
algorithms can be predicted for instance-specific schedules~\citep{amadini-tplp14a,lindauer-lion16a}.

Computing instance features can come with a large amount of overhead, and if the
objective is to minimize runtime, this overhead should be minimized.
For example, on industrial-like SAT instances,
computing some instance features can take more than half of the total time
budget.

For more
details on algorithm selection systems and the different approaches used in the
literature, we refer the interested reader to the surveys by
\cite{smith-miles-acmcs08a} and \cite{kotthoff-aim14a}.

\subsection{Evaluation of Algorithm Selection Systems}
The purpose of performing algorithm selection is to achieve performance better
than any individual algorithm could. In many cases,
overhead through the computation of the instance features used as input for the
machine learning models is incurred. This diminishes performance gains achieved through
selecting good algorithms and has to be taken into account for evaluating
algorithm selection systems.

To be able to assess the performance gain of algorithm selection systems,
two baselines are commonly compared against~\citep{xu-sat12a,lindauer-jair15a,ansotegui-aij16}:
\begin{inparaenum}[(i)]
\item the performance of the individual algorithm performing best on all training instances (called
\emph{single best solver (SBS)}), which denotes what can be achieved
without algorithm selection;
\item the performance of the \emph{virtual best solver (VBS)} (also called
oracle performance), which makes perfect decisions and chooses the
best-performing algorithm on each instance without any overhead. The
VBS corresponds to the overhead-free parallel portfolio that runs all algorithms
in parallel and terminates as soon as the first algorithm finishes.
\end{inparaenum}

The performance of the baselines and of any algorithm selection system varies
for different scenarios. We normalize the performance \mbox{$m_s = \sum_{\inst \in \insts} m (s(\inst), \inst)$} of an algorithm selection
system $s$ on a given scenario by the performance of the SBS
and VBS, as a cost to be minimized, and measure how much of the gap between the two it closed as follows:

\begin{equation}
    \hat{m}_s = \frac{m_s - m_{VBS}}{m_{SBS} - m_{VBS}}\label{eq:gap}
\end{equation}

where $0$ corresponds to perfect performance, equivalent to the VBS, and $1$
corresponds to the performance of the SBS.\footnote{In the 2017 competition, the
gap was defined such that $1$ corresponded to VBS and $0$ to SBS. For
consistency with the 2015 results, we use the metric as defined in Equation~\ref{eq:gap} here.} The
performance of an algorithm selection system will usually be between 0 and 1; if
it is larger than 1 it means that simply always selecting the SBS is a better strategy.

A common way of measuring runtime performance is penalized average runtime
(PAR10)~\citep{hutter-aij14a,lindauer-jair15a,ansotegui-aij16}: the average runtime
across all instances, where algorithms are run with a timeout and penalized with
a runtime ten times the timeout if they do not complete within the time limit.

\section{Competition Setups}
\label{sec:setup}

In this section, we discuss the setups of both competitions. Both competitions
were based on ASlib, with submissions required to read the ASlib format as
input.

\subsection{General Setup: ASlib}
\begin{figure}[t]
\tikzstyle{activity}=[rectangle, draw=black, rounded corners, text centered, text width=8em]
\tikzstyle{data}=[rectangle, draw=black, text centered, text width=8em]
\tikzstyle{myarrow}=[->, thick, draw=black]

\begin{tikzpicture}[node distance=5cm, thick]
	\node (Algo) [data] {Algorithm Portfolio $\algo \in \portfolio$};
	\node (Inst) [data, below of=Algo, node distance=5em] {Instances $\inst \in \insts$};
	
	\node (Perf) [data, right of=Algo] {Performance $m$ of each pair $\langle \algo, \inst \rangle_{\algo\in\portfolio, \inst\in\insts}$};
	\node (Descr) [data, above of=Perf, node distance=4.5em] {Scenario\\description};
	\node (FCost) [data, right of=Inst] {Cost to compute features for each $\inst\in\insts$};
	\node (Feats) [data, below of=FCost, node distance=4em] {Instance features for each $\inst\in\insts$};
	\node (Split) [data, below of=Feats, node distance=3.5em] {Train-Test splits of $\insts$};
	
	\node (Selector) [data, right of=Perf, node distance=5.4cm] {Build AS System};
	\node (Pred) [data, right of=Feats, node distance=5.4cm] {Predictions for Test Instances};
	
	\draw[myarrow] (Algo) -- (Perf);
	\draw[myarrow] (Inst.east) -- ($(Perf.west)+(0.0,-0.2)$);
	\draw[myarrow] (Inst.east) -- (Feats.west);
	\draw[myarrow] (Inst.east) -- (FCost.west);
	\draw[myarrow] (Inst.east) -- (Split.west);
	
	\draw[myarrow] ($(Perf.east)+(0.7,0)$) -- node[above] {$\insts_{Train}$} ($(Selector.west)+(0.0,+0.0)$);
	\draw[myarrow] ($(Feats.east)+(0.7,0)$) -- node[above] {$\insts_{Test}$} ($(Pred.west)+(0.0,+0.0)$);
	
    \draw[myarrow] (Selector) -- node[left,text centered, text width=4em] {Compare\\ \& Evaluate} (Pred);
	
	\draw[draw=black!60, dashed] ($(Descr)+(-2.5,+1.0)$) -- ($(Split)+(-2.5,-2.0)$);
	\draw[draw=black!60, dashed] ($(Descr)+(2.5,+1.0)$) -- ($(Split)+(2.5,-2.0)$);
	
	\node (ASlib) [below of=Split, node distance=4em, text=black!60] {ASlib Scenario Files};
	\node (Data) [left of=ASlib, node distance=12em, text=black!60] {Data Gathering};
	\node (Eva) [right of=ASlib, node distance=12em, text=black!60] {Evaluation of AS Systems};
	
\end{tikzpicture}

\caption{Illustration of ASlib.}\label{fig:aslib}
\end{figure}
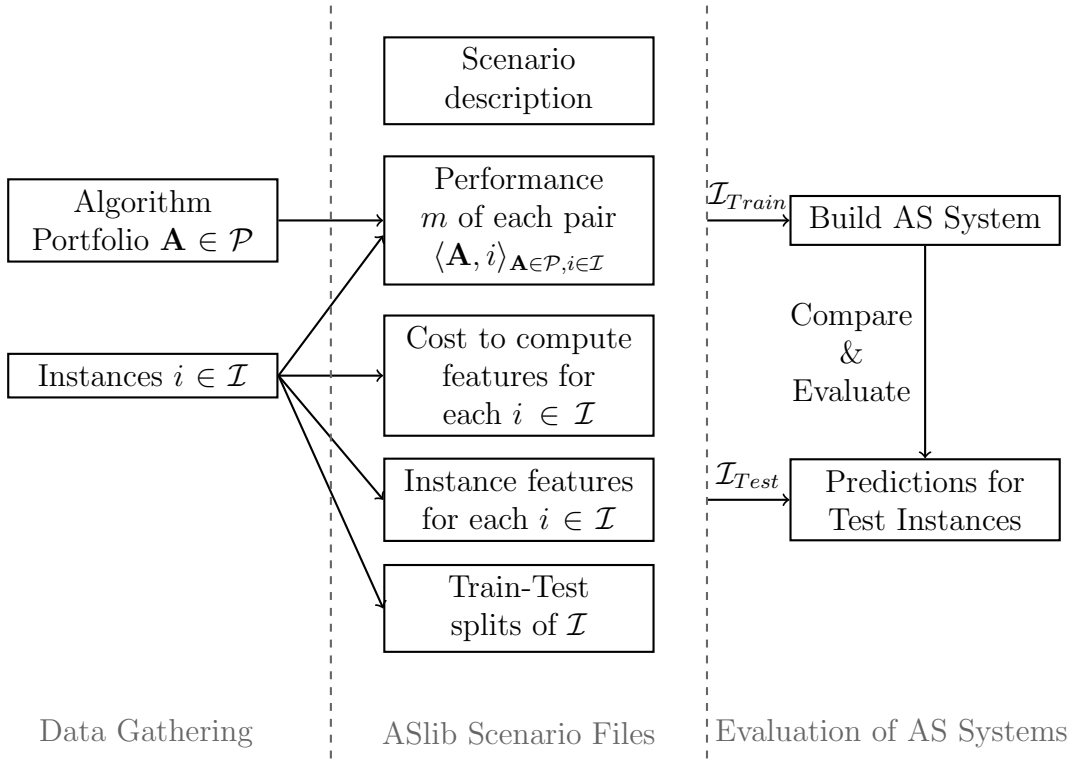

Figure~\ref{fig:aslib} shows the general structure of an ASlib \emph{scenario}~\citep{bischl-aij16a}.
ASlib scenarios contain pre-computed performance
values $m(\algo, \inst)$ for all algorithms in a portfolio $\algo \in
\portfolio$ on a set of training instances $\inst \in \insts$ (e.g., runtime
for SAT instances or accuracy for Machine Learning datasets). In addition, a set
of pre-computed instance features $\feat(\inst)$ are available for each instance, as well as
the time required to compute the feature values (the overhead). The corresponding task
description provides further information, e.g., runtime cutoff, grouping of
features, performance metric (runtime or solution quality) and indicates whether
the performance metric is to be maximized or minimized.
Finally, it contains a file describing the train-test splits. 
This file specifies which instances should be used for training the system
($\insts_{Train}$), and which should be used for evaluating the system ($\insts_{Test}$).

\subsection{Competition 2015}

In 2015, the competition asked for complete systems to be submitted which would
be trained and evaluated by the organizers. This way, the general applicability
of submissions was emphasized -- rather than doing well only with specific models
and after manual tweaks, submissions had to demonstrate that they can be used
off-the-shelf to produce algorithm selection models with good performance. For
this reason, submissions were required to be open source or free for academic
use.

The scenarios used in 2015 are shown in Table~\ref{tab:scens15}.
The competition used existing ASlib
scenarios that were known to the participants beforehand.
There was no secret test data in 2015; however, the splits
into training and testing data were not known to participants.
We note that these are all
runtime scenarios, reflecting what was available in ASlib at the time.

Submissions were allowed to specify the feature groups and a single pre-solver
for each ASlib scenario (a statically-defined algorithm to be run before any
feature computation to avoid overheads on easy instances), and required to
produce a list of the algorithms to run for each instance (each with an
associated timeout). The training time for a submission was limited to $12$ CPU
hours on each scenario; each submission had the same computational resources
available and was executed on the same hardware. AutoFolio was the
only submission that used the full $12$ hours.
The submissions were evaluated on $10$ different train-test splits,
to reduce the potential influence of randomness. We considered the three
metrics
mean PAR10 score, mean misclassification
penalty (the additional time that was required to solve an instance
compared to the best algorithm on that instance), and
number of instances solved within the timeout. The final score was the average
remaining gap $\hat{m}$ (Equation~\ref{eq:gap}) across these three metrics, the
$10$ train-test splits, and the scenarios.

\begin{table}[t!]
\begin{tabularx}{\textwidth}{Xl rrlr}
\hline
Scenario            & $|\algo|$ & $|\insts|$ & $|\feat|$  & Obj. & Factor\\
\hline
ASP-POTASSCO        & $11$      & $1294$     & $138$      & Time & $25$\\
CSP-2010            & $2$       & $2024$     & $17$       & Time & $10$\\
MAXSAT12-PMS        & $6$       & $876$      & $37$       & Time & $53$\\
CPMP-2013           & $4$       & $527$      & $22$       & Time & $31$\\
PROTEUS-2014        & $22$      & $4021$     & $198$      & Time & $413$\\
QBF-2011            & $5$       & $1368$     & $46$       & Time & $96$\\
SAT11-HAND          & $15$      & $296$      & $115$      & Time & $37$\\
SAT11-INDU          & $18$      & $300$      & $115$      & Time & $22$\\
SAT11-RAND          & $9$       & $600$      & $115$      & Time & $66$\\
SAT12-ALL           & $31$      & $1614$     & $115$      & Time & $30$\\
SAT12-HAND          & $31$      & $1167$     & $138$      & Time & $35$\\
SAT12-INDU          & $31$      & $767$      & $138$      & Time & $15$\\
SAT12-RAND          & $31$      & $1167$     & $138$      & Time & $12$\\
\hline
\end{tabularx}
\caption{Overview of algorithm selection scenarios used in 2015, showing
the number of algorithms $|\algo|$, the number of instances $|\insts|$, the
number of instance features $|\feat|$, the performance objective, and the
improvement factor of the virtual best solver (VBS) over the
single best solver $\left(m_\textit{SBS} / m_\textit{VBS}\right)$ without
considering instances on which all algorithms timed out.}
\label{tab:scens15}
\end{table}

\subsection{Competition 2017}

\begin{table}[t!]
\begin{tabularx}{\textwidth}{Xl rrrlr}
\hline
Scenario                & Alias   & $|\algo|$ & $|\insts|$ & $|\feat|$  & Obj. & Factor\\
\hline
BNSL-2016$^*$           & Bado    & $8$       & $1179$     & $86$    & Time      & $41$\\
CSP-Minizinc-Obj-2016   & Camilla & $8$       & $100$      & $95$    & Quality   & $1.7$\\
CSP-Minizinc-Time-2016  & Caren   & $8$       & $100$      & $95$    & Time      & $61$\\
MAXSAT-PMS-2016         & Magnus  & $19$      & $601$      & $37$    & Time      & $25$\\
MAXSAT-WPMS-2016        & Monty   & $18$      & $630$      & $37$    & Time      & $16$\\
MIP-2016                & Mira    & $5$       & $218$      & $143$   & Time      & $11$\\
OPENML-WEKA-2017        & Oberon  & $30$      & $105$      & $103$   & Quality   & $1.02$\\
QBF-2016                & Qill    & $24$      & $825$      & $46$    & Time      & $265$\\
SAT12-ALL$^*$           & Svea    & $31$      & $1614$     & $115$   & Time      & $30$\\
SAT03-16\_INDU          & Sora    & $10$      & $2000$     & $483$   & Time      & $13$\\
TTP-2016$^*$            & Titus   & $22$      & $9720$     & $50$    & Quality   & $1.04$\\
\hline
\end{tabularx}
\caption{Overview of algorithm selection scenarios used in 2017, showing the
alias in the competition, the number of algorithms $|\algo|$, the number of
instances $|\insts|$, the number of instance features $|\feat|$, the performance
objective, and the improvement factor of the virtual
best solver (VBS) over the single best solver $\left(m_\textit{SBS} /
m_\textit{VBS}\right)$ without considering instances on which all algorithms
timed out. Scenarios marked with an asterisk were available in ASlib before the
competition.}
\label{tab:scens17}
\end{table}

Compared to 2015, we changed the setup of the competition in 2017 with the following goals in mind:

\begin{enumerate}
    \item fewer restrictions on the submissions regarding computational
    resources and licensing;
    \item better scaling of the organizational overhead to more submissions, in
    particular not having to run each submission manually;
    \item more flexible schedules for computing features and running algorithms; and
    \item a more diverse set of algorithm selection scenarios, including new scenarios.
\end{enumerate}

To achieve the first and second goal, the participants did not submit their systems
directly, but only the predictions made by their system for new test instances
(using a single train-test split). Thus, also submissions from closed-source
systems were possible, although all participants made their submissions
open-source in the end. We provided full information, including algorithms'
performances, for a set of training instances, but only the feature values for
the test instances to submitters. Participants could invest as many
computational resources as they wanted to compute their predictions.
While this may give an advantage to participants who have access to large
amounts of computational resources,
such a competition is typically won through better ideas
and not through better computational resources.
To facilitate easy submission of results, we did not run multiple train-test
splits as in 2015. We be briefly investigated the effects of this in Section~\ref{sec:randomness}.
We note that this setup is quite common in other machine learning competitions,
e.g., the Kaggle competitions~\citep{Carpenter2011}.

To support more complex algorithm selection approaches, the submitted
predictions were allowed to be an arbitrary sequence of algorithms with timeouts
and interleaved feature computations. Thus, any combination of these two
components was possible (e.g., complex pre-solving schedules with interleaved
feature computation). Complex pre-solving schedules were used by
most submissions for scenarios with runtime as performance metric.

We collected several new algorithm selection benchmarks from different domains;
$8$ out of the $11$ used scenarios were completely new and not disclosed to
participants before the competition (see Table~\ref{tab:scens17}). 
We obfuscated the instance and algorithm names such that
the participants were not able to easily recognize existing scenarios. 

To show the impact of
algorithm selection on the state of the art in different domains, we focused the
search for new scenarios on recent competitions for CSP, MAXSAT, MIP, QBF, and
SAT. Additionally, we developed an open-source Python tool that connects to
OpenML~\citep{Vanschoren2014} and converts a Machine Learning study into an
ASlib scenario.\footnote{See \url{https://github.com/openml/openml-aslib}.}
To ensure diversity of the scenarios with respect to the application
domains, we selected at most two scenarios from each domain
to avoid any bias introduced by focusing on a single domain. In the 2015
competition, most of the scenarios came from SAT, which skewed the evaluation in
favor of that. Finally, we also considered scenarios with solution quality as
performance metric (instead of runtime) for the first time. The new scenarios
were added to ASlib after the competition; thus the competition was not only
enabled by ASlib, but furthers its expansion.

For a detailed description of the competition setup in 2017, we refer the interested reader to \cite{Lindauer17}.

\section{Results}
\label{sec:results}

We now discuss the results of both competitions.

\subsection{Competition 2015}

The competition received a total of 8 submissions from 4 different groups of
researchers comprising 15 people. Participants were based in 4 different
countries on 2 continents. \ref{app:sys15} provides
an overview of all submissions.

\begin{table}[t]
\centering
\begin{tabularx}{\textwidth}{lXrr}
  \toprule
  Rank & System & \multicolumn{2}{c}{Avg. Gap}\\
  \cline{3-4}
       &   & \multicolumn{1}{c}{All} & \multicolumn{1}{c}{PAR10}\\
  \midrule
  1st & zilla \dotfill               & 0.366 & 0.344 \\
  2nd & zillafolio \dotfill          & 0.370 & 0.341 \\
  ooc & AutoFolio-48 \dotfill        & 0.375 & 0.334 \\
  3rd & AutoFolio \dotfill           & 0.390 & 0.341 \\
  ooc & LLAMA-regrPairs \dotfill     & 0.395 & 0.375 \\
  4th & ASAP\_RF \dotfill            & 0.416 & 0.377 \\
  5th & ASAP\_kNN \dotfill           & 0.423 & 0.387 \\
  ooc & LLAMA-regr \dotfill          & 0.425 & 0.407 \\
  6th & flexfolio-schedules \dotfill & 0.442 & 0.395 \\
  7th & sunny \dotfill               & 0.482 & 0.461 \\
  8th & sunny-presolv \dotfill       & 0.484 & 0.467 \\
  \bottomrule
\end{tabularx}
\caption{Results in 2015 with some system running out of competition (ooc). The
average gap is aggregated across all scenarios according to Equation~\ref{eq:gap}.}
\label{tab:ranks}
\end{table}

Table~\ref{tab:ranks} shows the final ranking. The zilla system is the overall
winner, although the first- and second-placed entries are very close. All
systems perform well on average, closing more than half of the gap between
virtual and single best solver.
Additionally, we show the normalized PAR10 score for comparison to the 2017 results,
where only the PAR10 metric was used.
Detailed results of all metrics (PAR10,
misclassification penalty, and solved) are presented in~\ref{sec:results_icon}.

\begin{figure}[t]
    \centering
    \includegraphics[width=\textwidth]{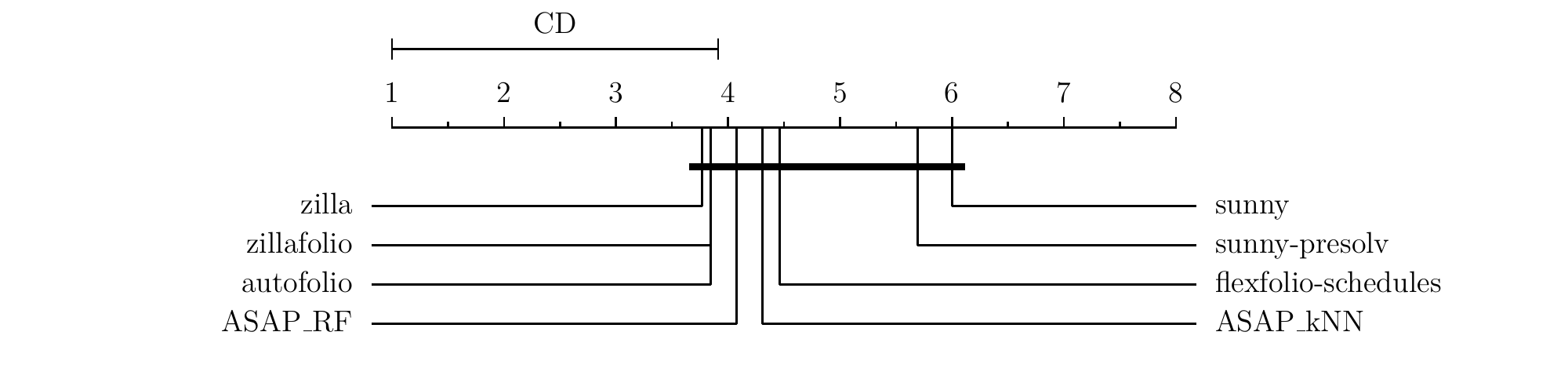}
    \caption{Critical distance plots with Nemenyi Test on the `All' scores (average across normalized scores based
    on PAR10, misclassification penalty, and number of solved instances) of the participants of the 2015 competition.
    If two submissions are connected by a thick line, there was not enough statistical evidence that
    their performances are significantly different.}
    \label{fig:icon_nemenyi_ALL}
\end{figure}

\begin{figure}[t]
    \centering
    \includegraphics[width=\textwidth]{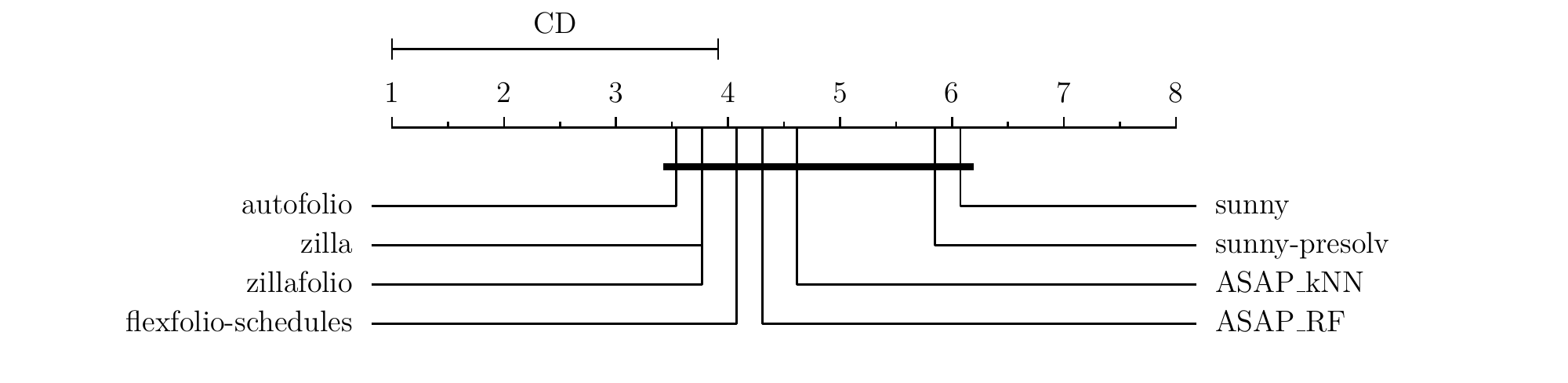}
    \caption{Critical distance plots with Nemenyi Test on the PAR10 scores of the participants of the 2015 competition.}
    \label{fig:icon_nemenyi_PAR10}
\end{figure}

For comparison, we show three additional systems. Autofolio-48 is identical to
Autofolio (a submitted algorithm selector that searches over different selection approaches
and their hyperparameter settings~\citep{lindauer-jair15a}),
but was allowed $48$ hours training time (four times the default)
to assess the impact of additional tuning of hyperparameters.
LLAMA-regrPairs and
LLAMA-regr are simple approaches based on the LLAMA algorithm selection toolkit~\citep{kotthoff-tech13a}.\footnote{
Both LLAMA approaches use regression models to
predict the performance of each algorithm individually and
for each pair of algorithms to predict their performance difference.
Both approaches did not use pre-solvers and feature selection, both selected only a single algorithm, and
their hyperparameters were not tuned.}
The relatively small difference between AutoFolio and AutoFolio-48
shows that allowing more training time does not increase performance
significantly. The good ranking of the two simple LLAMA models shows that
reasonable performance can be achieved even with simple off-the-shelf approaches
without customization or tuning.
Figure~\ref{fig:icon_nemenyi_ALL} (combined scores) and Figure~\ref{fig:icon_nemenyi_PAR10} (PAR10 scores)
show critical distance plots on the average ranks of the submissions. 
According to the Friedman test with post-hoc Nemenyi test, there is no 
statistically significant difference between any of the submissions.

More detailed results can be found in~\cite{kotthoff-iconas15}.

\subsection{Competition 2017}

In 2017, there were $8$ submissions from $4$ groups.
Similar to 2015, participants were based in 4 different
countries on 2 continents. While most of the submissions came from
participants of the 2015 competition, there were also submissions by researchers
who did not participate in 2015.

\begin{table}[t]
    \centering
    \begin{tabularx}{\textwidth}{l X rr}
    \toprule
    Rank &  System                   & Avg. Gap    & Avg. Rank\\
    \midrule
    1st & ASAP.v2 \dotfill           & $0.38$     & $2.6$\\
    2nd & ASAP.v3 \dotfill           & $0.40$     & $2.8$\\
    3rd & Sunny-fkvar \dotfill       & $0.43$     & $2.7$\\
    4th & Sunny-autok \dotfill       & $0.57$     & $3.9$\\
    ooc  & $^*$Zilla(fixed version) \dotfill & $0.57$     & N/A \\
    5th & $^*$Zilla \dotfill         & $0.93$     & $5.3$\\
    6th & $^*$Zilla(dyn) \dotfill    & $0.96$     & $5.4$\\
    7th & AS-RF \dotfill             & $2.10$     & $6.1$\\
    8th & AS-ASL \dotfill            & $2.51$     & $7.2$\\
    \bottomrule
    \end{tabularx}
    \caption{Results in 2017 with some system running out of competition (ooc).
    The average gap is aggregated across all scenarios according to
    Equation~\ref{eq:gap}.}
    \label{tab:results_oasc}
\end{table}

Table~\ref{tab:results_oasc} shows the results in terms of the gap metric (see
Equation~\ref{eq:gap}) based on PAR10, as well as the ranks; detailed results
are in Table~\ref{tab:results_oasc:app} (\ref{sec:results_oasc}). The
competition was won by ASAP.v2, which obtained the highest scores on the gap metric
both in terms of the average over all datasets, and the average rank across all
scenarios. Both ASAP systems clearly outperformed all other participants on the
quality scenarios. However, Sunny-fkvar did best on the runtime scenarios, followed by
ASAP.v2.

\begin{figure}[t]
    \centering
    \includegraphics[width=\textwidth]{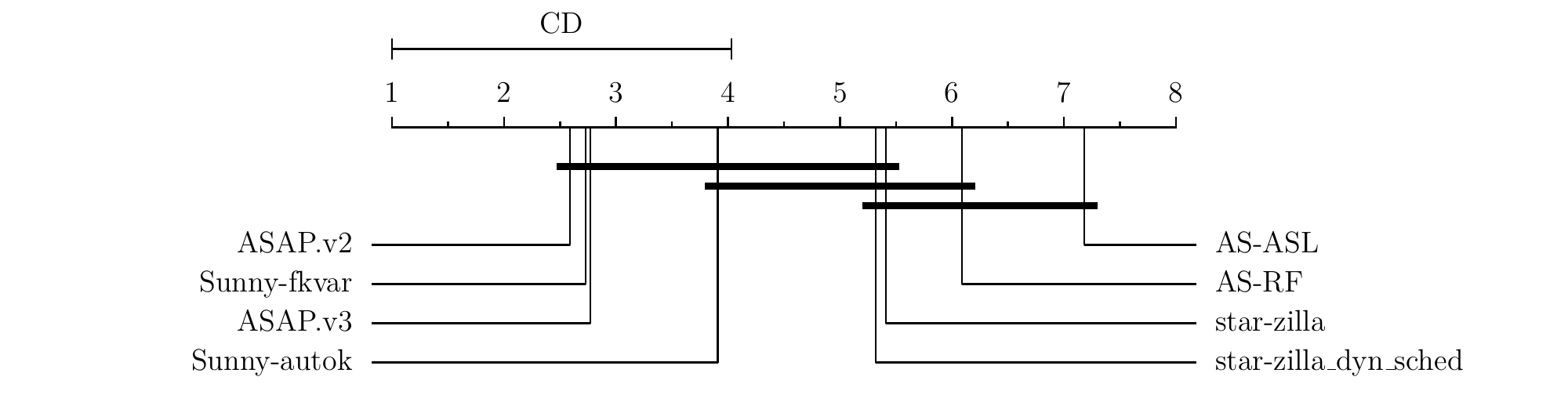}
    \caption{Critical distance plots with Nemenyi Test on the PAR10 scores of the participants in 2017.}
    \label{fig:oasc_nemenyi}
\end{figure}

Figure~\ref{fig:oasc_nemenyi} shows critical distance plots on the average ranks
of the submissions. 
There is no statistically significant difference between the best six
submissions, but the difference to the worst submissions is statistically
significant.

\section{Open Challenges and Insights}
\label{sec:insights}

In this section, we discuss insights and open challenges indicated by the
results of the competitions.

\subsection{Progress from 2015 to 2017}

The progress of algorithm selection as a field from 2015 to 2017 seems to be
rather small. In terms of the remaining gap between virtual best and single best
solver, the results were nearly the same (the best system in 2015 achieved about
$33\%$ in terms of PAR10, and the best system in 2017 about $38\%$). On
the only scenario used in both competitions (SAT12-ALL), the performance stayed
nearly constant. Nevertheless, the competition in 2017 was more challenging because
of the new and more diverse scenarios. While the community succeeded in coming up
with more challenging problems, there appears to be room for more innovative
solutions.

\subsection{Statistical Significance}
Figures~\ref{fig:icon_nemenyi_ALL},~\ref{fig:icon_nemenyi_PAR10} 
and~\ref{fig:oasc_nemenyi} show ranked plots, with the critical distance required
according to the Friedman with post-hoc Nemenyi test to assert statistical 
significant difference between multiple systems~\citep{Demsar2006}. 
In the 2015 competition, none of the differences between the submitted systems
were
statistical significant, whereas in the 2017 competition only some differences
where statistical significant.

Failure to detect a significant difference does not
imply that there is no such difference: the statistical 
tests are based on a relatively low number of samples and thus have limited
power.

Even though the statistical significance results should be interpreted 
with care, the critical difference plots are still informative. They show, e.g.,
that the systems submitted in the 2015 challenge were closer together 
(ranked approximately between $3.5$ and $6$) 
than the systems submitted in 2017 (ranked approximately between $2.5$ and
$7$).

\subsection{Robustness of Algorithm Selection Systems}\label{sec:robust}

As the results of both competitions show, choosing one of the state-of-the-art
algorithm selection systems is still a much better choice than simply always
using the single best algorithm. However, as different algorithm
selection systems have different strengths, we are now confronted with a
meta-algorithm selection problem -- selecting the best algorithm selection
approach for the task at hand. For example, while the best submission in 2017
achieved $38\%$ gap between SBS and VBS remaining, the virtual best selector
over the portfolio of all submissions would have achieved $29\%$. An open
challenge is to develop such a meta-algorithm selection system, or a single
algorithm selection system that performs well across a wide range of scenarios.

One step in this direction is the per-scenario customization of the systems, e.g., by
using hyperparameter optimization methods~\citep{Gonard17,Liu17,Cameron17},
per-scenario model selection~\citep{Malone17}, or even per-scenario
selection of the general approach combined with hyperparameter
optimization~\citep{lindauer-jair15a}. However, as the
results show, more fine-tuning of an algorithm selection system does not
always result in a better-performing system. In 2015, giving much more time to
Autofolio resulted in only a very minor performance improvement, and in 2017
ASAP.v2 performed better than its refined successor ASAP.v3.

In addition to the general observations above, we note the following points regarding robustness of the submissions:

\begin{itemize}
    \item zilla performed very well on SAT scenarios (average rank: $1.4$) but
    only mediocre on other domains (average rank: $6.5$ out of $8$ submissions)
    in 2015;
    \item ASAP won in 2017, but sunny-fkvar performed better on runtime
    scenarios;
    \item both CSP scenarios in 2017 were very similar (same algorithm
    portfolio, same instances, same instance features) but the performance
    metric was changed (one scenario with runtime and one scenario with solution
    quality). On the runtime scenario, Sunny-fkvar performed very well, but on
    the quality scenario ASAP.v3/2 performed much better.
\end{itemize}

\subsection{Impact of Randomness\label{sec:randomness}}
One of the main differences between the 2015 and 2017 challenges was that in
2015, the submissions were evaluated on 10 cross-validation
splits to determine the final ranking, whereas in 2017, only a single training-test split
was used. While this greatly reduced the effort for the competition organizers,
it increased the risk of a particular submission with randomized components
getting lucky.

In general, our expectation for the performance of a submission is that it does
not depend on randomness much, i.e., its performance does not vary significantly
across different test sets or random seeds. On the other hand, as we observed in
Section~\ref{sec:robust}, achieving good performance across multiple scenarios
is an issue.

\begin{figure}
  \begin{center}
    \includegraphics[width=\textwidth]{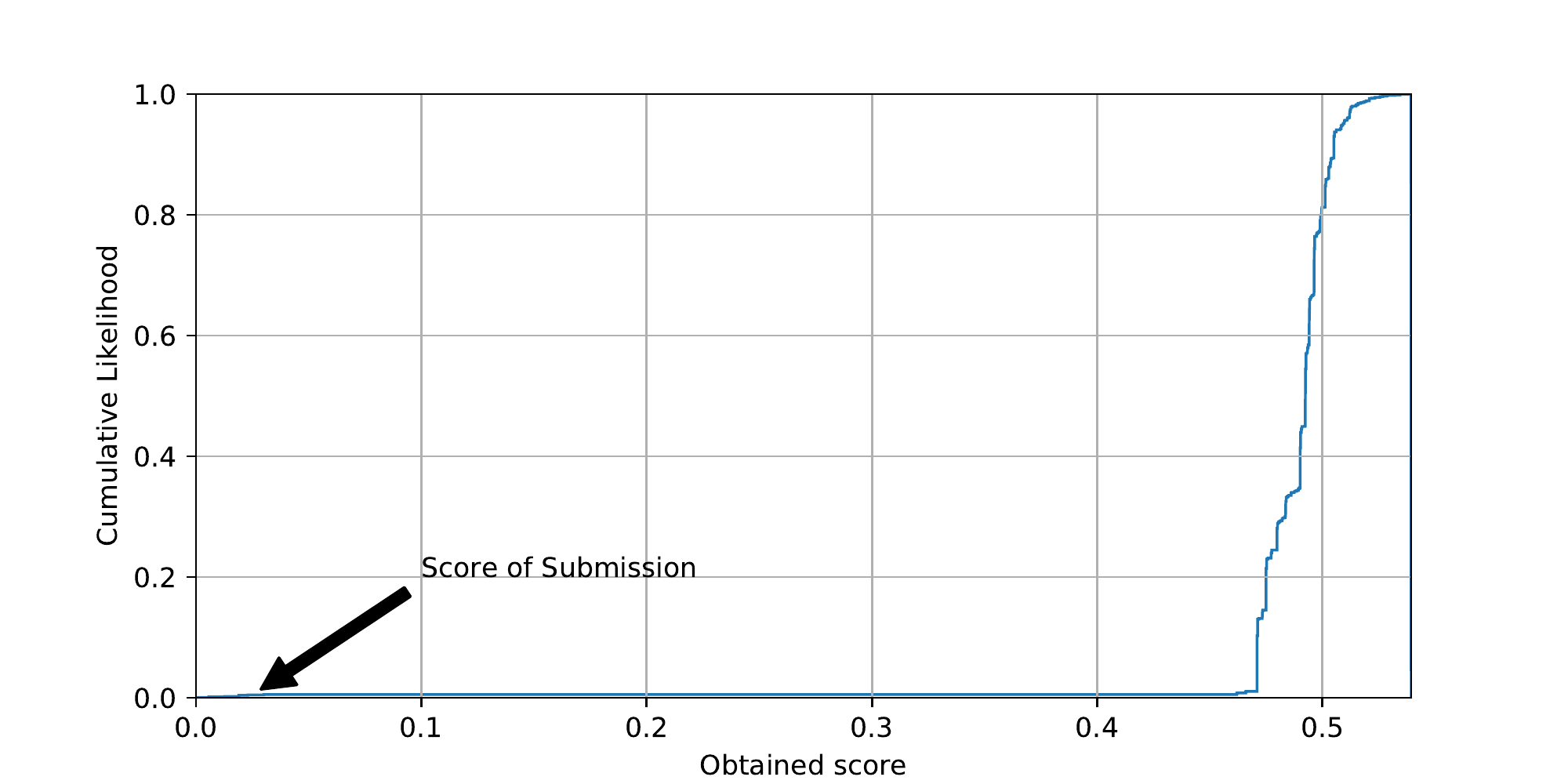}
    \caption{Cumulative distribution function of the closed gap of ASAP.v2 on
    CSP-Minizinc-Obj-2016, across 1500 random seeds. The plot shows that the
    actual obtained score ($0.025$) has a probability of $0.466\%$.
    \label{fig:asap_camilla}}
  \end{center}
\end{figure}

To determine the effect of randomness on performance, we ran the competition
winner, ASAP.v2, with different random seeds on the CSP-Minizinc-Obj-2016
(Camilla) scenario, where it performed particularly well.
Figure~\ref{fig:asap_camilla} shows the cumulative distribution function of the
performance across different random seeds. The probability of ASAP.v2 performing
as good or better than it did is very low, suggesting that it did choose a lucky
random seed.

This result demonstrates the importance of evaluating algorithm selection
systems across multiple random seeds, or multiple test sets. If we replace
ASAP's obtained score with the median score of the CDF shown in
Figure~\ref{fig:asap_camilla}, it would have ranked at third place.

\subsection{Hyperparameter Optimization}
All systems submitted to either of the competitions leverage a machine learning
model that predicts the performance of algorithms.
It is well known that hyperparameter optimization is important to get
well-performing machine learning models
(see, e.g., \cite{snoek-nips12a,thornton-kdd13a,Rijn2018b}).
Nevertheless, not all submissions optimized hyperparameters, e.g., the winner in
2017 ASAP.v2~\citep{Gonard17} used the default hyperparameters of its random
forest. Given previous results by \cite{lindauer-jair15a}, we would expect that
adding hyperparameter optimization to recent algorithm selection systems will
further boost their performances.

\subsection{Handling of Quality Scenarios}

ASlib distinguishes between two types of scenarios: \emph{runtime scenarios} and
\emph{quality scenarios}. In runtime scenarios, the goal is to minimize the time
the selected algorithm requires to solve an instances (e.g., SAT, ASP), whereas in
quality scenarios the goal is to find the algorithm that obtains the highest
score or lowest error according to some metric (e.g., plan quality in AI
planning or prediction error in Machine Learning). In the current version of
ASlib, the most important difference between the two scenario types is that for
runtime scenario a schedule of different algorithms can be provided, whereas for
quality scenarios only a single algorithm. The reason for this limitation is
that ASlib does not contain information on intermediate solution qualities of
any-time algorithms (e.g., the solution quality after half the timeout). For the
same reason, the cost of feature computation cannot be considered for quality
scenarios -- it is unknown how much additional quality could be achieved in
the time required for feature computation. This setup is common in algorithm
selection methods for machine learning (meta-learning). Intermediate solutions
and the time at which they were obtained could enable schedules for quality
scenarios and analyzing trade-offs between
obtaining a better solution quality by expending more resources or switching to
another algorithm. For example, the MiniZinc
Challenge~\citep{stuckey-aim14a} started to record these information in 2017.
Future versions of ASlib will consider addressing this limitation.

\subsection{Challenging Scenarios}

\begin{table}[t!]
\footnotesize
\begin{tabularx}{\textwidth}{X rr}
\toprule
Scenario            & Median rem. gap 	& Best rem. gap\\
\midrule
2015&&\\
\midrule
ASP-POTASSCO\dotfill         	& $0.31$ 	& $0.28$\\
CSP-2010\dotfill	            & $0.23$ 	& $0.14$\\
MAXSAT12-PMS\dotfill         	& $0.18$ 	& $0.14$\\
CPMP-2013\dotfill            	& $0.35$ 	& $0.29$\\
PROTEUS-2014\dotfill         	& $0.16$ 	& $0.05$\\
QBF-2011\dotfill             	& $0.15$ 	& $0.09$\\
SAT11-HAND\dotfill           	& $0.34$ 	& $0.30$\\
SAT11-INDU\dotfill           	& $1.00$ 	& $\mathbf{0.87}$\\
SAT11-RAND\dotfill           	& $0.08$ 	& $0.04$\\
SAT12-ALL\dotfill            	& $0.38$ 	& $0.27$\\
SAT12-HAND\dotfill           	& $0.32$ 	& $0.25$\\
SAT12-INDU\dotfill           	& $0.90$ 	& $0.59$\\
SAT12-RAND\dotfill           	& $1.00$ 	& $\mathbf{0.77}$\\
\midrule
Average							& $0.41$	& $0.31$ \\
\toprule
2017&&\\
\midrule
BNSL-2016\dotfill            	& $0.25$	& $0.15$ \\
CSP-Minizinc-Obj-2016\dotfill    & $1.59$	& $0.02$ \\
CSP-Minizinc-Time-2016\dotfill   & $0.41$	& $0.05$ \\
MAXSAT-PMS-2016\dotfill          & $0.49$	& $0.41$ \\
MAXSAT-WPMS-2016\dotfill         & $0.51$	& $0.08$ \\
MIP-2016\dotfill                 & $0.56$	& $0.49$ \\
OPENML-WEKA-2017\dotfill         & $1.0$	& $\mathbf{0.78}$ \\
QBF-2016\dotfill                 & $0.43$	& $0.15$ \\
SAT12-ALL\dotfill       			& $0.42$	& $0.31$ \\
SAT03-16\_INDU\dotfill           & $0.77$	& $\mathbf{0.65}$ \\
TTP-2016$^*$\dotfill             & $0.33$	& $0.15$\\
\midrule
Average					& $0.61$	& $0.30$ \\
\bottomrule
\end{tabularx}
\caption{Average remaining gap and the best remaining gap across all submissions
for all scenarios. The \textbf{bold} scenarios are particularly challenging.}
\label{tab:chall}
\end{table}

On average, algorithm selection systems perform well and the best
systems had a remaining gap between the single best and virtual best solver of
only $38\%$ in 2017. However, some of the scenarios are harder than others
for algorithm selection. Table~\ref{tab:chall} shows the median and best
performance of all submissions on all scenarios. To identify challenging
scenarios, we studied the best-performing submission on each scenario and
compared the remaining gap with the average remaining gap over all scenarios.
In 2015, \emph{SAT12-RAND} and \emph{SAT11-INDU} were particularly challenging,
and in 2017 \emph{OPENML-WEKA-2017} and \emph{SAT03-16\_INDU}.

\begin{figure}[tb!]
  \begin{center}
    \includegraphics[width=\textwidth]{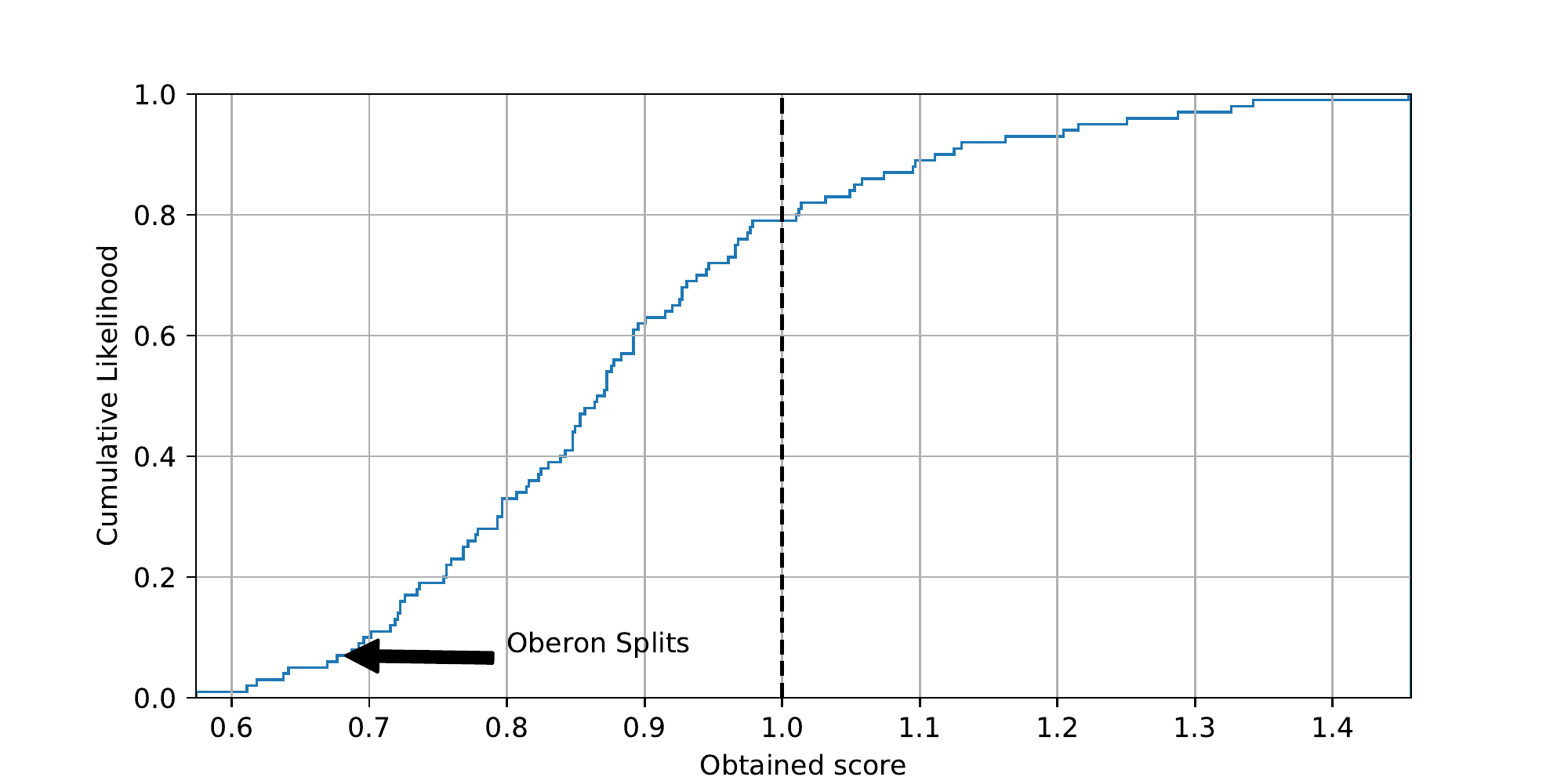}
    \caption{Cumulative distribution function of the obtained gap-remaining score
    of a random forest regressor (a single model trained to predict for all classifiers, 64 trees) on $100$ randomly sampled $33\%$ holdout sets of the OPENML-WEKA-2017 scenario.
	  The dashed line indicates the performance of the single best solver; the score on the actual splits as presented in Oberon was $0.675$. \label{fig:oberon_splits}}
  \end{center}
\end{figure}

\begin{description}
  \item[SAT12-RAND] was a challenging scenario in 2015 and most of the
  participating systems performed not better than the single best solver on it,
  although the VBS has a 12-fold speedup over the single best solver.
  The main reason is probably that not only the SAT instances
  considered in this scenario are randomly generated but also most of the
  best-performing solvers are stochastic local search solvers which are highly
  randomized. The data in this scenario was obtained from single runs of each
  algorithm, which introduces strong noise.
  After the competition in 2015,
  \cite{cameron-ijca16a} showed that in such noisy scenarios, the performance of
  the virtual best solver is often overestimated.
  Thus, we do not recommend to study algorithm selection on \emph{SAT12-RAND} at
  this moment and plan to remove \emph{SAT12-RAND} in the next ASlib release.
  \item[SAT11-INDU] was a hard scenario in 2015; in particular it was hard
  for systems that selected schedules per
  instance (such as Sunny). Applying schedules on these industrial-like
  instances is quite hard because even the single best solver has an average
  PAR10 score of $8030$ (with a timeout of $5000$ seconds) to solve an
  instance; thus, allocating a fraction of the total available resources to an algorithm on
  this scenario is often not a good idea (also shown by \cite{hoos-tplp14b}).
  \item[SAT03-16\_INDU] was a challenging scenario for the participants in $2017$.
  It is mainly an extension of a previously-used scenario
  called \emph{SAT12-INDU}. Zilla was one of the best submissions in 2015 on \emph{SAT12-INDU}
  with a remaining gap of roughly $61\%$; however
  in 2017 on \emph{SAT03-16\_INDU}, zilla had a remaining gap of $83\%$. Similar observations apply to ASAP.
  \emph{SAT03-16\_INDU} could be much harder than \emph{SAT12-INDU} because of
  the smaller number of algorithms ($31\to 10$),
  the larger number of instances ($767 \to 2000$)
  or the larger number of instance features ($138 \to 483$).
  \item[OPENML-WEKA-2017] was a new scenario in the 2017 competition and
  appeared to be very challenging, as six out of eight submissions performed
  almost equal to or worse than the single best solver ($\geq 95\% $ remaining gap). This scenario featured algorithm
  selection for machine learning problems (cf.\ meta-learning~\citep{Brazdil2008}).
  The objective was to select the best machine learning algorithm from a
  selection of WEKA algorithms~\citep{Hall2009}, based on simple characteristics
  of the dataset (i.e., meta-features). The scenario was introduced by
  \cite{Rijn2016}[Chapter~6]. We verified empirically that
  \begin{inparaenum}[(i)]
    \item there is a learnable concept in this scenario, and
    \item the chosen holdout set was sufficiently similar to the training data
  \end{inparaenum}
  by evaluating a simple baseline algorithm selector (a regression approach using a random forest as model).
  The experimental setup and results are presented in
  Figure~\ref{fig:oberon_splits}.
  It is indeed a challenging scenario; on half of the sampled holdout sets,
  our baseline was unable to close the gap by more than $10\%$.
  In $18\%$ of the holdout sets, the baseline performed worse than the SBS.
  However, our simple baseline achieved $67.5\%$ remaining gap on the holdout set used in the competition (compared to the best submission Sunny-fkvar with $78\%$).
\end{description}

\section{Conclusions}

In this paper, we discussed the setup and results of two algorithm
selection competitions. These competitions allow the community to objectively
compare different systems and assess their benefits. They confirmed that
per-instance algorithm selection can substantially improve the state of the art
in many areas of AI. For example, the virtual best solver obtains on average a
31.8 fold speedup over the single best solver on the runtime scenarios from
2017. While the submissions fell short of this perfect performance, they did
achieve significant improvements.

Perhaps more importantly, the competitions highlighted challenges for the
community in a field that has been well-established for more than a decade. We
identified several challenging scenarios on which the recent algorithm selection
systems do not perform well. Furthermore, there is no system that performs well
on all types of scenarios -- a meta-algorithm selection problem is very much
relevant in practice and warrants further research. The competitions also
highlighted restrictions in the current version of ASlib, which enabled the
competitions, that need to be addressed in future work.


\section*{Acknowledgments}
Marius Lindauer acknowledges funding by the DFG (German Research
Foundation) under Emmy Noether grant HU 1900/2-1.

\section*{References}
\bibliographystyle{elsarticle-harv}
\bibliography{strings,lib,local,proc,oasc2017}

\newpage

\appendix

\section{Submitted Systems in 2015}
\label{app:sys15}


\begin{itemize}
  \item \emph{ASAP} based on random forests (RF) and $k$-nearest neighbor (kNN) as selection models combine pre-solving schedule and per-instance algorithm selection by training both jointly~\citep{gonard-meta16}.
  \item \emph{AutoFolio} combines several algorithm selection approaches in a single systems and uses algorithm configuration~\citep{hutter-lion11a} to search for the best approach and its hyperparameter settings for the scenario at hand.
  \item \emph{Sunny} selects an algorithm schedule on a per-instance base~\citep{amadini-tplp14a}. The time assigned to each algorithm is proportional to the number of solved instances in the neighborhood in the feature space with respect to the instance at hand.
  \item \emph{Zilla} is the newest version of SATzilla~\citep{xu-jair08a,xu-rcra11a} which uses pair-wise, cost-sensitive random forests combined with pre-solving schedules.
  \item \emph{ZillaFolio} is a combination of Zilla and AutoFolio by evaluating both approaches on the training set and using the better approach for generating the predictions for the test set.
\end{itemize}

\section{Technical Evaluation Details in 2015}

The evaluation was performed as follows. For each scenario, 10 bootstrap
samples of the entire data were used to create 10 different train/test splits.
No stratification was used. The training part was left unmodified. For the test
part, algorithm performances were set to 0 and runstatus to ``ok'' for all
algorithms and all instances -- the ASlib specification requires algorithm
performance data to be part of a scenario.

There was a time limit of 12 hours for the training phase. Systems that exceeded
this limit were disqualified. The time limit was chosen for practical reasons,
to make it possible to evaluate the submissions with reasonable resource
requirements.

For systems that specified a pre-solver, the instances that were solved by the
pre-solver within the specified time were removed from the training set. If a
subset of features was specified, only these features (and only the costs
associated with these features) were left in both training and test set, with
all other feature values removed.

Each system was trained on each train scenario and predicted on each test
scenario. In total, 130 evaluations (10 for each of the 13 scenarios) per
submitted system were performed. The total CPU time spent was
4685.11 hours on 8-core Xeon E5-2640 CPUs.

Each system was evaluated in terms of mean PAR10 score, mean misclassification
penalty (the additional time that was required to solve an instance because an
algorithm that was not the best was chosen; the difference to the VBS), and mean
number of instances solved for each of the 130 evaluations on each scenario and
split. These are the same performance measures used in ASlib, and enable a
direct comparison.

The final score of a submission group (i.e.\ a system submitted for different
ASlib scenarios) was computed as the average score over all ASlib scenarios. For
scenarios for which no system belonging to the group was submitted, the
performance of the single best algorithm was assumed.

\section{Submitted Systems in 2017}
\label{app:sys17}

\begin{itemize}
    \item \cite{Gonard17} submitted \emph{ASAP.v2 and ASAP.v3}~\citep{gonard-meta16}. ASAP combines pre-solving schedules and per-instance algorithm selection by training both jointly. The main difference between ASAP.v2 and ASAP.v3 is that ASAP.v2 used a pre-solving schedule with a fixed length of 3, whereas ASAP.v3 optimized the schedule length between 1 and 4 on a per-scenario base.
    \item \cite{Malone17} submitted \emph{AS-RF and AS-ASL}~\citep{malone-mlj18a}. It also combines pre-solving schedules and per-instance algorithm selection, whereas the selection model is a two-level stacking model with the first level being regression models to predict the performance of each algorithm and the second level combines these performance predictions in a multi-class model to obtain a selected algorithm. AS-RF uses random forest and AS-ASL used auto-sklearn~\citep{feurer-nips2015a} to obtain a machine learning model.
    \item \cite{Liu17} submitted \emph{Sunny (autok and fkvar)}~\citep{amadini-tplp14a}. Sunny selects per-instance algorithm schedules with the goal of minimizing the number of possible timeouts. Sunny-autok optimized the neighborhood size on a per-scenario base~\citep{lindauer-lion16a} and Sunny.fkvar additionally also applied greedy forward selection for instance feature subset selection.
    \item \cite{Cameron17} submitted \emph{*Zilla (vanilla and dynamic)}, the successor of SATzilla~\citep{xu-jair08a,xu-rcra11a}. *Zilla also combines per-solving schedules and pre-instance algorithm selection but based on pair-wise weighted random forest models. The dynamic version of *Zilla additionally uses the trained random forest to extract a per-instance algorithm schedule.\footnote{*Zilla had a critical bug and the results were strongly degraded because of it. The authors of *Zilla submitted fixed results after the official deadline but before the test data and the results were announced. We list the fixed results of *Zilla; but these are not officially part of the competition.}
\end{itemize}

\clearpage
\section{Detailed Results 2015 competition}
\label{sec:results_icon}

\begin{table}[ht]
\scriptsize
\begin{tabularx}{\textwidth}{X rrrrr}
\toprule
scenario & zilla & zillafolio & autofolio & flexfolio-schedules \\
\midrule
ASP-POTASSCO & $537$ ($5.0$) & $516$ ($1.0$) & $525$ ($3.0$) & $527$ ($4.0$) \\
CSP-2010 & $6582$ ($4.0$) & $6549$ ($2.0$) & $6621$ ($7.0$) & $6573$ ($3.0$) \\
MAXSAT12-PMS & $3524$ ($6.0$) & $3598$ ($8.0$) & $3559$ ($7.0$) & $3375$ ($1.0$) \\
PREMARSHALLING-ASTAR-2013 & $2599$ ($5.0$) & $2722$ ($7.0$) & $2482$ ($4.0$) & $2054$ ($2.0$) \\
PROTEUS-2014 & $5324$ ($7.0$) & $5070$ ($5.0$) & $5057$ ($4.0$) & $4435$ ($1.0$) \\
QBF-2011 & $9339$ ($7.0$) & $9366$ ($8.0$) & $9177$ ($6.0$) & $8653$ ($1.0$) \\
SAT11-HAND & $17436$ ($3.0$) & $17130$ ($1.0$) & $17746$ ($6.0$) & $17560$ ($4.0$) \\
SAT11-INDU & $13418$ ($3.0$) & $13768$ ($4.0$) & $13314$ ($1.0$) & $14560$ ($6.0$) \\
SAT11-RAND & $9495$ ($2.0$) & $9731$ ($3.0$) & $9428$ ($1.0$) & $10339$ ($8.0$) \\
SAT12-ALL & $964$ ($1.0$) & $1100$ ($3.0$) & $1066$ ($2.0$) & $1436$ ($6.0$) \\
SAT12-HAND & $4370$ ($2.0$) & $4432$ ($4.0$) & $4303$ ($1.0$) & $4602$ ($6.0$) \\
SAT12-INDU & $2754$ ($3.0$) & $2680$ ($1.0$) & $2688$ ($2.0$) & $2972$ ($4.0$) \\
SAT12-RAND & $3139$ ($1.0$) & $3146$ ($2.0$) & $3160$ ($3.0$) & $3240$ ($7.0$) \\
\midrule
Average & $6114$ ($3.8$) & $6139$ ($3.8$) & $6087$ ($3.6$) & $6179$ ($4.1$) \\
\bottomrule
\toprule
scenario & ASAP\_RF & ASAP\_kNN & sunny & sunny-presolv \\
\midrule
ASP-POTASSCO & $517$ ($2.0$) & $554$ ($7.0$) & $575$ ($8.0$) & $547$ ($6.0$) \\
CSP-2010 & $6516$ ($1.0$) & $6601$ ($5.0$) & $6615$ ($6.0$) & $6704$ ($8.0$) \\
MAXSAT12-PMS & $3421$ ($3.0$) & $3395$ ($2.0$) & $3465$ ($4.0$) & $3521$ ($5.0$) \\
PREMARSHALLING-ASTAR-2013 & $2660$ ($6.0$) & $2830$ ($8.0$) & $2151$ ($3.0$) & $1979$ ($1.0$) \\
PROTEUS-2014 & $5169$ ($6.0$) & $5338$ ($8.0$) & $4866$ ($3.0$) & $4798$ ($2.0$) \\
QBF-2011 & $8793$ ($2.0$) & $8813$ ($3.0$) & $8907$ ($4.0$) & $9044$ ($5.0$) \\
SAT11-HAND & $17581$ ($5.0$) & $17289$ ($2.0$) & $19130$ ($7.0$) & $19238$ ($8.0$) \\
SAT11-INDU & $13858$ ($5.0$) & $13359$ ($2.0$) & $14681$ ($7.0$) & $15160$ ($8.0$) \\
SAT11-RAND & $10018$ ($6.0$) & $9795$ ($4.0$) & $10212$ ($7.0$) & $9973$ ($5.0$) \\
SAT12-ALL & $1201$ ($5.0$) & $1181$ ($4.0$) & $1579$ ($7.0$) & $1661$ ($8.0$) \\
SAT12-HAND & $4434$ ($5.0$) & $4395$ ($3.0$) & $4823$ ($7.0$) & $4875$ ($8.0$) \\
SAT12-INDU & $3005$ ($6.0$) & $2974$ ($5.0$) & $3201$ ($8.0$) & $3173$ ($7.0$) \\
SAT12-RAND & $3211$ ($4.0$) & $3239$ ($6.0$) & $3263$ ($8.0$) & $3222$ ($5.0$) \\
\midrule
Average & $6183$ ($4.3$) & $6136$ ($4.5$) & $6421$ ($6.1$) & $6453$ ($5.8$) \\
\bottomrule
\end{tabularx}
\caption{Original results of the PAR10 scores of the 2015 competition.\label{tab:results_icon:par10}}
\end{table}

\begin{table}[ht]
\scriptsize
\begin{tabularx}{\textwidth}{X rrrrr} \\
\toprule
scenario & zilla & zillafolio & autofolio & flexfolio-schedules \\
\midrule
ASP-POTASSCO & $22$ ($5.0$) & $21$ ($2.0$) & $22$ ($3.0$) & $24$ ($7.0$) \\
CSP-2010 & $14$ ($2.0$) & $11$ ($1.0$) & $28$ ($7.0$) & $23$ ($6.0$) \\
MAXSAT12-PMS & $38$ ($2.0$) & $42$ ($4.0$) & $177$ ($8.0$) & $41$ ($3.0$) \\
PREMARSHALLING-ASTAR-2013 & $323$ ($5.0$) & $336$ ($7.0$) & $330$ ($6.0$) & $307$ ($4.0$) \\
PROTEUS-2014 & $482$ ($8.0$) & $470$ ($7.0$) & $470$ ($6.0$) & $70$ ($1.0$) \\
QBF-2011 & $192$ ($6.0$) & $194$ ($8.0$) & $182$ ($5.0$) & $133$ ($3.0$) \\
SAT11-HAND & $462$ ($3.0$) & $406$ ($1.0$) & $486$ ($5.0$) & $514$ ($6.0$) \\
SAT11-INDU & $615$ ($2.0$) & $639$ ($3.0$) & $574$ ($1.0$) & $779$ ($8.0$) \\
SAT11-RAND & $70$ ($3.0$) & $65$ ($2.0$) & $62$ ($1.0$) & $448$ ($8.0$) \\
SAT12-ALL & $95$ ($1.0$) & $111$ ($3.0$) & $103$ ($2.0$) & $211$ ($8.0$) \\
SAT12-HAND & $75$ ($1.0$) & $82$ ($3.0$) & $77$ ($2.0$) & $160$ ($8.0$) \\
SAT12-INDU & $87$ ($1.0$) & $100$ ($2.0$) & $103$ ($3.0$) & $139$ ($5.0$) \\
SAT12-RAND & $39$ ($1.0$) & $40$ ($2.0$) & $49$ ($4.0$) & $58$ ($5.0$) \\
\midrule
Average & $194$ ($3.1$) & $194$ ($3.5$) & $205$ ($4.1$) & $224$ ($5.5$) \\
\bottomrule
\toprule
scenario & ASAP\_RF & ASAP\_kNN & sunny & sunny-presolv \\
\midrule
ASP-POTASSCO & $23$ ($6.0$) & $25$ ($8.0$) & $20$ ($1.0$) & $22$ ($4.0$) \\
CSP-2010 & $14$ ($3.0$) & $21$ ($4.0$) & $22$ ($5.0$) & $39$ ($8.0$) \\
MAXSAT12-PMS & $45$ ($6.0$) & $43$ ($5.0$) & $31$ ($1.0$) & $57$ ($7.0$) \\
PREMARSHALLING-ASTAR-2013 & $271$ ($1.0$) & $275$ ($2.0$) & $338$ ($8.0$) & $296$ ($3.0$) \\
PROTEUS-2014 & $235$ ($4.0$) & $249$ ($5.0$) & $224$ ($3.0$) & $136$ ($2.0$) \\
QBF-2011 & $98$ ($2.0$) & $80$ ($1.0$) & $181$ ($4.0$) & $194$ ($7.0$) \\
SAT11-HAND & $466$ ($4.0$) & $431$ ($2.0$) & $634$ ($8.0$) & $618$ ($7.0$) \\
SAT11-INDU & $736$ ($6.0$) & $673$ ($4.0$) & $701$ ($5.0$) & $772$ ($7.0$) \\
SAT11-RAND & $124$ ($6.0$) & $129$ ($7.0$) & $122$ ($5.0$) & $85$ ($4.0$) \\
SAT12-ALL & $157$ ($5.0$) & $153$ ($4.0$) & $182$ ($6.0$) & $183$ ($7.0$) \\
SAT12-HAND & $114$ ($5.0$) & $102$ ($4.0$) & $124$ ($6.0$) & $129$ ($7.0$) \\
SAT12-INDU & $160$ ($8.0$) & $154$ ($7.0$) & $154$ ($6.0$) & $138$ ($4.0$) \\
SAT12-RAND & $60$ ($7.0$) & $67$ ($8.0$) & $58$ ($6.0$) & $45$ ($3.0$) \\
\midrule
Average & $193$ ($4.8$) & $185$ ($4.7$) & $215$ ($4.9$) & $209$ ($5.4$) \\
\bottomrule
\end{tabularx}
\caption{Original results of the misclassification penalty scores of the 2015 competition.\label{tab:results_icon:mcp}}
\end{table}

\begin{table}[ht]
\scriptsize
\begin{tabularx}{\textwidth}{X rrrrr} \\
\toprule
scenario & zilla & zillafolio & autofolio & flexfolio-schedules \\
\midrule
ASP-POTASSCO & $0.915$ ($5.0$) & $0.919$ ($2.0$) & $0.917$ ($4.0$) & $0.918$ ($3.0$) \\
CSP-2010 & $0.870$ ($4.0$) & $0.871$ ($2.0$) & $0.870$ ($6.0$) & $0.870$ ($3.0$) \\
MAXSAT12-PMS & $0.834$ ($7.0$) & $0.830$ ($8.0$) & $0.840$ ($4.0$) & $0.842$ ($1.0$) \\
PREMARSHALLING-ASTAR-2013 & $0.937$ ($5.0$) & $0.933$ ($6.0$) & $0.940$ ($4.0$) & $0.953$ ($2.0$) \\
PROTEUS-2014 & $0.863$ ($6.0$) & $0.871$ ($3.0$) & $0.871$ ($2.0$) & $0.878$ ($1.0$) \\
QBF-2011 & $0.745$ ($7.0$) & $0.744$ ($8.0$) & $0.750$ ($6.0$) & $0.765$ ($1.0$) \\
SAT11-HAND & $0.659$ ($3.0$) & $0.665$ ($1.0$) & $0.653$ ($6.0$) & $0.658$ ($4.0$) \\
SAT11-INDU & $0.741$ ($3.0$) & $0.734$ ($5.0$) & $0.742$ ($2.0$) & $0.719$ ($6.0$) \\
SAT11-RAND & $0.815$ ($2.0$) & $0.809$ ($3.0$) & $0.816$ ($1.0$) & $0.804$ ($6.0$) \\
SAT12-ALL & $0.930$ ($1.0$) & $0.918$ ($3.0$) & $0.921$ ($2.0$) & $0.897$ ($6.0$) \\
SAT12-HAND & $0.643$ ($3.0$) & $0.638$ ($5.0$) & $0.649$ ($1.0$) & $0.629$ ($6.0$) \\
SAT12-INDU & $0.779$ ($3.0$) & $0.787$ ($1.0$) & $0.787$ ($2.0$) & $0.764$ ($5.0$) \\
SAT12-RAND & $0.742$ ($1.0$) & $0.742$ ($2.0$) & $0.741$ ($3.0$) & $0.735$ ($7.0$) \\
\midrule
Average & $0.806$ ($3.8$) & $0.805$ ($3.8$) & $0.807$ ($3.3$) & $0.802$ ($3.9$) \\
\bottomrule
\toprule
scenario & ASAP\_RF & ASAP\_kNN & sunny & sunny-presolv \\
\midrule \\
ASP-POTASSCO & $0.919$ ($1.0$) & $0.913$ ($7.0$) & $0.908$ ($8.0$) & $0.913$ ($6.0$) \\
CSP-2010 & $0.872$ ($1.0$) & $0.870$ ($5.0$) & $0.870$ ($7.0$) & $0.868$ ($8.0$) \\
MAXSAT12-PMS & $0.840$ ($3.0$) & $0.841$ ($2.0$) & $0.837$ ($5.0$) & $0.835$ ($6.0$) \\
PREMARSHALLING-ASTAR-2013 & $0.933$ ($7.0$) & $0.928$ ($8.0$) & $0.951$ ($3.0$) & $0.955$ ($1.0$) \\
PROTEUS-2014 & $0.861$ ($7.0$) & $0.856$ ($8.0$) & $0.870$ ($4.0$) & $0.869$ ($5.0$) \\
QBF-2011 & $0.759$ ($2.0$) & $0.758$ ($4.0$) & $0.758$ ($3.0$) & $0.754$ ($5.0$) \\
SAT11-HAND & $0.656$ ($5.0$) & $0.662$ ($2.0$) & $0.625$ ($7.0$) & $0.623$ ($8.0$) \\
SAT11-INDU & $0.734$ ($4.0$) & $0.744$ ($1.0$) & $0.715$ ($7.0$) & $0.706$ ($8.0$) \\
SAT11-RAND & $0.804$ ($7.0$) & $0.809$ ($4.0$) & $0.800$ ($8.0$) & $0.804$ ($5.0$) \\
SAT12-ALL & $0.913$ ($5.0$) & $0.915$ ($4.0$) & $0.881$ ($7.0$) & $0.873$ ($8.0$) \\
SAT12-HAND & $0.640$ ($4.0$) & $0.643$ ($2.0$) & $0.605$ ($7.0$) & $0.601$ ($8.0$) \\
SAT12-INDU & $0.763$ ($6.0$) & $0.765$ ($4.0$) & $0.744$ ($8.0$) & $0.745$ ($7.0$) \\
SAT12-RAND & $0.738$ ($4.0$) & $0.736$ ($5.0$) & $0.733$ ($8.0$) & $0.735$ ($6.0$) \\
Average & $0.802$ ($4.3$) & $0.803$ ($4.3$) & $0.792$ ($6.3$) & $0.791$ ($6.2$) \\
\bottomrule
\end{tabularx}
\caption{Original results of the solved scores of the 2015 competition.\label{tab:results_icon:solved}}
\end{table}

\clearpage

\section{Detailed Results 2017 competition}
\label{sec:results_oasc}

\begin{table}[ht]
\footnotesize
\begin{tabularx}{\textwidth}{X rrrrr} \\
\toprule
scenario & ASAP.v2 & ASAP.v3 & Sunny-fkvar & Sunny-autok \\
\midrule
Bado & $0.239$ ($4.0$) & $0.192$ ($3.0$) & $0.153$ ($1.0$) & $0.252$ ($5.0$) \\
Camilla & $0.025$ ($1.5$) & $0.025$ ($1.5$) & $0.894$ ($3.0$) & $1.475$ ($4.0$) \\
Caren & $0.412$ ($5.0$) & $0.410$ ($4.0$) & $0.055$ ($1.0$) & $0.217$ ($2.0$) \\
Magnus & $0.492$ ($4.0$) & $0.494$ ($5.0$) & $0.419$ ($3.0$) & $0.498$ ($6.0$) \\
Mira & $0.495$ ($2.0$) & $0.491$ ($1.0$) & $0.568$ ($4.0$) & $1.014$ ($6.0$) \\
Monty & $0.167$ ($2.0$) & $0.237$ ($3.0$) & $0.090$ ($1.0$) & $0.368$ ($4.0$) \\
Oberon & $0.950$ ($3.5$) & $0.950$ ($3.5$) & $0.787$ ($1.0$) & $0.877$ ($2.0$) \\
Quill & $0.302$ ($2.0$) & $0.420$ ($3.0$) & $0.431$ ($4.0$) & $0.150$ ($1.0$) \\
Sora & $0.650$ ($1.0$) & $0.775$ ($4.0$) & $0.821$ ($5.0$) & $0.827$ ($6.0$) \\
Svea & $0.324$ ($2.0$) & $0.312$ ($1.0$) & $0.342$ ($3.0$) & $0.421$ ($4.0$) \\
Titus & $0.154$ ($1.5$) & $0.154$ ($1.5$) & $0.201$ ($4.0$) & $0.195$ ($3.0$) \\
\midrule
Average & $0.383$ ($2.6$) & $0.405$ ($2.8$) & $0.433$ ($2.7$) & $0.572$ ($3.9$) \\
\bottomrule
\toprule
scenario & star-zilla\_dyn\_sched & star-zilla & AS-RF & AS-ASL \\
\midrule
Bado & $0.516$ ($8.0$) & $0.293$ ($6.0$) & $0.164$ ($2.0$) & $0.319$ ($7.0$) \\
Camilla & $3.218$ ($7.5$) & $3.218$ ($7.5$) & $1.974$ ($5.0$) & $2.289$ ($6.0$) \\
Caren & $0.223$ ($3.0$) & $1.001$ ($6.0$) & $1.659$ ($7.0$) & $2.068$ ($8.0$) \\
Magnus & $0.410$ ($1.0$) & $0.417$ ($2.0$) & $2.012$ ($7.0$) & $2.013$ ($8.0$) \\
Mira & $2.337$ ($8.0$) & $0.967$ ($5.0$) & $0.505$ ($3.0$) & $1.407$ ($7.0$) \\
Monty & $0.513$ ($5.0$) & $0.827$ ($6.0$) & $8.482$ ($8.0$) & $7.973$ ($7.0$) \\
Oberon & $1.000$ ($5.5$) & $1.000$ ($5.5$) & $3.798$ ($7.0$) & $7.233$ ($8.0$) \\
Quill & $0.541$ ($5.0$) & $0.692$ ($6.0$) & $1.328$ ($8.0$) & $1.299$ ($7.0$) \\
Sora & $0.687$ ($2.5$) & $0.687$ ($2.5$) & $1.135$ ($7.0$) & $1.383$ ($8.0$) \\
Svea & $0.829$ ($7.5$) & $0.829$ ($7.5$) & $0.543$ ($5.0$) & $0.561$ ($6.0$) \\
Titus & $0.335$ ($5.5$) & $0.335$ ($5.5$) & $1.535$ ($8.0$) & $1.113$ ($7.0$) \\
\midrule
Average & $0.964$ ($5.3$) & $0.933$ ($5.4$) & $2.103$ ($6.1$) & $2.514$ ($7.2$) \\
\bottomrule
\end{tabularx}
\caption{Original results of the 2017 competition -- score of 0 refers to the virtual best solver
and 1 to the single best solver.\label{tab:results_oasc:app}}
\end{table}
\clearpage

\end{document}